\documentclass{article}
\usepackage{graphicx} 
\usepackage{natbib}
\usepackage{amsmath, amssymb}
\usepackage{subcaption} 
\usepackage{xcolor} 
\usepackage{hyperref}       
\hypersetup{
	colorlinks=true,
	linkcolor=blue,
	filecolor=magenta,      
	urlcolor=cyan,
	citecolor=blue,
}
\usepackage{enumitem}
\usepackage{amsthm}
\usepackage{multirow}
\usepackage{booktabs}
\usepackage{caption}
\usepackage{geometry}
\usepackage{mathtools}
\usepackage{cleveref}
\newcommand{\norm}[1]{\left\lVert#1\right\rVert}

\newcommand{\Span}[1]{\mathrm{span}\left\langle #1 \right\rangle}
\newtheorem{phenomenon}{Phenomenon}

\crefname{phenomenon}{phenomenon}{phenomena}

\newif\ifcomments
\commentstrue 

\ifcomments
\newcommand{\sps}[1]{{\color{olive}[SPS: #1]}}
\else
\newcommand{\sps}[1]{}
\fi

\title{Accelerating Neural Network Training Along Sharp and Flat Directions}
\author{Daniyar Zakarin\thanks{Correspondence to dzakarin@ethz.ch, contact@sidakpal.com}, Sidak Pal Singh
        }

\begin{document}

\maketitle

\begin{abstract}
Recent work has highlighted a surprising alignment between gradients and the top eigenspace of the Hessian—termed the \textit{Dominant subspace}—during neural network training. Concurrently, there has been growing interest in the distinct roles of sharp and flat directions in the Hessian spectrum. In this work, we study \textit{Bulk}-SGD, a variant of SGD that restricts updates to the orthogonal complement of the Dominant subspace. Through ablation studies, we characterize the stability properties of \textit{Bulk}-SGD and identify critical hyperparameters that govern its behavior. We show that updates along the \textit{Bulk subspace}, corresponding to flatter directions in the loss landscape, can accelerate convergence but may compromise stability. To balance these effects, we introduce interpolated gradient methods that unify \textit{SGD}, \textit{Dom}-SGD, and \textit{Bulk}-SGD. Finally, we empirically connect this subspace decomposition to the \textit{Generalized Gauss–Newton} and \textit{Functional Hessian} terms, showing that curvature energy is largely concentrated in the Dominant subspace. Our findings suggest a principled approach to designing curvature-aware optimizers.

\end{abstract}

\section{Introduction \& Related Work}

Stochastic Gradient Descent (SGD) remains the cornerstone of optimization in deep learning. Any advancement that improves its convergence speed or ability to find better solutions is of substantial practical interest. A variety of first-order methods, including Momentum (\cite{pmlr-v28-sutskever13}) and Adam (\cite{kingma2017adammethodstochasticoptimization}), have been proposed to address limitations of vanilla SGD. These algorithms are often motivated by insights into the geometry of neural network loss landscapes—particularly the need to efficiently traverse sharp and flat regions.

Recent empirical investigations have illuminated a surprising structure in the Hessian of deep networks: it is typically dominated by a small number of large eigenvalues, suggesting a low-rank curvature profile \cite{sagun2017eigenvalueshessiandeeplearning, sagun2018empiricalanalysishessianoverparametrized, pmlr-v97-ghorbani19b, papyan2019measurementsthreelevelhierarchicalstructure}. \cite{gurari2018gradientdescenthappenstiny} observed that the gradient vectors during training tend to align with the top-k eigenspace of the Hessian for k-class classification tasks. They hypothesized that learning predominantly occurs within this dominant subspace.

However, this hypothesis was challenged in \cite{song2025doessgdreallyhappen}, where authors performed controlled experiments to test its validity. Their results show that restricting updates to the dominant subspace fails to decrease the training loss beyond a certain point. In contrast, updates constrained to the orthogonal complement of this subspace achieve similar performance. This finding suggests that the observed alignment is not causally linked to effective optimization but is instead a byproduct of stochastic noise inherent in SGD.

These insights offer both theoretical and practical implications. From a mechanistic standpoint, they suggest that much of the optimization trajectory may be spent moving in directions informed more by noise than by meaningful signal. This opens the door to the possibility that constraining updates to be orthogonal to high-curvature directions—effectively filtering out noisy components—could enable more stable and potentially larger steps. Such a strategy may accelerate convergence or help escape suboptimal basins by focusing descent along flatter directions in the landscape.

\subsection{Description of the Main Results}

We summarize our main findings as follows:

\begin{itemize}
    \item In \cref{section:ablation_study}, we conduct a series of ablation studies to evaluate \textit{Bulk}-SGD as a standalone training algorithm. We assess its sensitivity to key hyperparameters including the learning rate, the dimension of the \textit{Dominant subspace}, and the \textit{Hessian holdout size}. We analyze how the behaviors described in \cref{phenom:dom,phenom:bulk} vary when switching from standard SGD to \textit{Bulk}-SGD at different training steps. Through these ablation studies identify and provide an explanation for an instability problem of \textit{Bulk}-SGD. 
    
    \item In \cref{section:acceleration}, we explore how the flatness of the \textit{Bulk subspace} can be leveraged to accelerate neural network training. Through controlled experiments, we demonstrate that \textit{Bulk}-SGD achieves faster convergence than classical SGD. In \cref{phenom:acceleration_vs_stability}, we hypothesize distinct roles for the \textit{Dominant} and \textit{Bulk} components of the gradient—namely, that the \textit{Bulk} component promotes acceleration, while the \textit{Dominant} component contributes to stability. We further investigate the connection between these dynamics and generalization in \cref{phenom:acceleration_vs_generalization}. To address the trade-off between acceleration and stability, we introduce an interpolated gradient method that unifies \textit{SGD}, \textit{Bulk}-SGD, and \textit{Dom}-SGD. We study how different interpolation regimes affect training loss and test accuracy.
    
    \item In \cref{section:hessian_decomp}, we investigate the relationship between the \textit{Dominant} and \textit{Bulk subspaces} and the decomposition of the Hessian into the \textit{Generalized Gauss–Newton} and \textit{Functional Hessian} terms. We find that the energies of the full Hessian and the \textit{Generalized Gauss–Newton} component align closely with the \textit{Dominant subspace}, while the \textit{Functional Hessian} exhibits little alignment. 
\end{itemize}

\section{Background}

Let $\theta \in \mathbb{R}^p$ denote the parameters of a neural network trained with a twice-differentiable loss function $\mathcal{L} : \mathbb{R}^p \rightarrow \mathbb{R}$. Following the notation of \cite{song2025doessgdreallyhappen}, we consider the spectrum of the Hessian $\nabla^2 \mathcal{L}(\theta)$ at a given point in the parameter space. Let $\lambda_1(\theta) \geq \lambda_2(\theta) \geq \dots \geq \lambda_p(\theta)$ denote its ordered eigenvalues, with corresponding orthonormal eigenvectors $u_1(\theta), u_2(\theta), \dots, u_p(\theta)$.

We define the \emph{Dominant subspace} to be the span of the top-$k$ eigenvectors:
\[
    S_k(\theta) = \mathrm{Span}\left\{u_1(\theta), u_2(\theta), \dots, u_k(\theta)\right\}.
\]
In practice, $k$ is typically chosen to match the number of output classes in a classification task.

The orthogonal projector on this dominant subspace is denoted:
\[
    P_k(\theta) \coloneqq \sum_{i = 1}^k u_i(\theta) u_i(\theta)^\top.
\]

Let $g = \nabla \mathcal{L}(\theta)$ denote the gradient of the loss. We decompose it into components aligned with the \emph{Dominant} and complementary \emph{Bulk} subspaces:
\[
    g = P_k(\theta) g + P_k^\perp(\theta) g, \quad \text{where} \quad P_k^\perp(\theta) \coloneqq I - P_k(\theta).
\]

\cite{gurari2018gradientdescenthappenstiny} made a notable empirical observation about this decomposition:

\begin{phenomenon}
\textbf{SGD gradients tend to align with the dominant subspace.}

Define the \emph{alignment coefficient} as
\[
    \chi_k(\theta) \coloneqq \frac{\|P_k(\theta) g\|_2}{\|g\|_2}, \quad \text{where} \quad g = \nabla \mathcal{L}(\theta).
\]
They observe that, after a few initial steps of training with stochastic gradient descent (SGD), the alignment $\chi_k(\theta_t)$ rapidly increases and stays near 1 throughout training.
\label{phenom:alignment}
\end{phenomenon}

To test whether learning truly occurs along the dominant subspace, \cite{song2025doessgdreallyhappen} propose two variants of gradient descent:

\vspace{0.5em}
\begin{itemize}
    \item \textit{Dom}-SGD: $\theta_{t+1} \leftarrow \theta_t - \eta P_k(\theta_t) g_t$
    \item \textit{Bulk}-SGD: $\theta_{t+1} \leftarrow \theta_t - \eta P_k^\perp(\theta_t) g_t$
\end{itemize}
\vspace{0.5em}

Here, $g_t = \nabla \mathcal{L}(\theta_t)$ is the gradient at time $t$, and $\eta$ is the learning rate.

The following phenomena are observed:

\begin{phenomenon}
\textbf{\textit{Dom}-SGD is insufficient to reduce training loss.}

After training a model with standard SGD, switching to \textit{Dom}-SGD halts further decrease in training loss. This suggests that, despite the apparent alignment, the \textit{Dominant subspace} does not support continued learning.
\label{phenom:dom}
\end{phenomenon}

\begin{phenomenon}
\textbf{\textit{Bulk}-SGD retains full training dynamics.}

When switching to \textit{Bulk}-SGD, the training loss exhibits the same dynamics as the vanilla SGD. 
\label{phenom:bulk}
\end{phenomenon}

\section{Ablation Study}
\subsection{Bulk-SGD as a Standalone Optimizer}
\label{section:ablation_study}

\begin{table}[ht]
\centering
\begin{tabular}{cc|cccc|c}
\toprule
$\eta$ & $l$ & $k=1$ & $k=5$ & $k=10$ & $k=20$ & SGD \\
\midrule
0.01 & 200  & 39.59 & 39.09 & 38.98 & 37.70 & \multirow{3}{*}{\textbf{39.67}} \\
0.01 & 500  & 39.55 & 39.00 & 38.86 & 38.40 & \\
0.01 & 5000 & 39.46 & 38.02 & 38.55 & 37.57 & \\
\midrule
0.05 & 200  & 39.58 & 39.00 & \textcolor{red}{28.95} & \textcolor{red}{28.63} & \multirow{3}{*}{\textbf{39.67}} \\
0.05 & 500  & 39.38 & \textcolor{red}{35.07} & 38.12 & \textcolor{red}{18.05} & \\
0.05 & 5000 & 39.60 & 38.45 & 38.38 & \textcolor{red}{35.88} & \\
\midrule
0.1  & 200  & \textcolor{red}{30.49} & 40.71 & \textcolor{red}{40.02} & \textcolor{red}{19.92} & \multirow{3}{*}{40.84} \\
0.1  & 500  & \textbf{41.12} & 39.74 & \textcolor{red}{29.74} & \textcolor{red}{29.40} & \\
0.1  & 5000 & 40.81 & 39.83 & \textcolor{red}{39.53} & \textcolor{red}{32.52} & \\
\bottomrule
\end{tabular}
\caption{Test Accuracy on CIFAR10-5k using MLP (ReLU) with varying learning rates, Dominant subspace dimensions, and Hessian holdout sizes. The red values indicate the runs with the diverging loss. The best accuracy for each learning rate is indicated in bold.}
\label{table:ablation_mlp}
\end{table}

\begin{table}[ht]
\centering
\caption{Test Accuracy on CIFAR10-5k using CNN with varying learning rates, dominant subspace dimensions, and Hessian holdout sizes. The best accuracy for each learning rate is indicated in bold.}
\begin{tabular}{cc|cccc|c}
\toprule
LR & Param & 1 & 5 & 10 & 20 & SGD \\
\midrule
0.001 & 200 & 41.69 & 40.73 & 42.28 & 42.18 & 41.98 \\
0.001 & 500 & 41.64 & 39.39 & \textbf{42.43} & 42.29 & \\
\midrule
0.01  & 200 & 54.35 & 37.83 & 53.88 & 54.17 & \textbf{54.83} \\
0.01  & 500 & 54.37 & 34.61 & 53.52 & 52.76 & \\
\bottomrule
\end{tabular}
\label{table:ablation_cnn}
\end{table}

We evaluate \textit{Bulk}-SGD as a standalone training algorithm, with particular focus on its sensitivity to key hyperparameters. Unlike the setup in \ref{phenom:dom}, \ref{phenom:bulk}, we initialize training directly with \textit{Bulk}-SGD, foregoing any warmup with standard SGD.

\textbf{Projection update frequency.} Computing the projection onto the bulk subspace requires the leading eigenvectors of the Hessian. Following \cite{gurari2018gradientdescenthappenstiny}, we leverage the empirical observation that these eigenvectors are relatively stable across short time scales. In our experiments, we found that updating the projection up to every 100 steps has negligible effect on learning. For efficiency, we recompute the projection every 10 steps in the results reported in Table~\ref{table:ablation_mlp} and \ref{table:ablation_cnn}.

\textbf{Ablation.} To better understand the role of each hyperparameter, we perform a systematic ablation over:
\begin{itemize}
    \item dominant subspace dimension $k$,
    \item learning rate $\eta$, and
    \item the size $l$ of the dataset used to estimate the Hessian.
\end{itemize}
To avoid information leakage, the Hessian is computed on a holdout set sampled from the training data prior to training. We refer to $l$ as the \emph{Hessian holdout size}. Results are summarized in Tables~\ref{table:ablation_mlp} and \ref{table:ablation_cnn}.

In the absence of divergence, \textit{Bulk}-SGD often attains accuracy on par with or better than SGD, consistent with Phenomenon~\ref{phenom:bulk}.

\textbf{Stability.} Despite promising results, \textit{Bulk}-SGD frequently exhibits instability at larger learning rates, with training losses diverging early. We analyze this phenomenon in detail in Section~\ref{section:instability}.

We find that increasing $k$ typically degrades test accuracy in the MLP setting, revealing a weak negative correlation. In contrast, the Hessian holdout size $l$ has little observable impact on final performance.

\subsection{Switch Point Sensitivity of \textit{Dom}/\textit{Bulk}-SGD}
\begin{figure}[ht]
    \centering
    \begin{subfigure}[b]{0.32\linewidth}
        \centering
        \includegraphics[width=\linewidth]{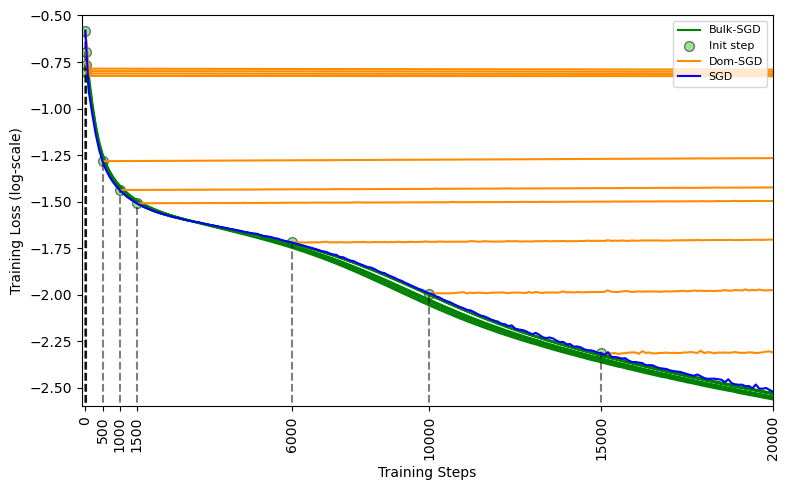}
        \caption{Training Loss}
        \label{fig:sub1}
    \end{subfigure}
    \hfill
    \begin{subfigure}[b]{0.32\linewidth}
        \centering
        \includegraphics[width=\linewidth]{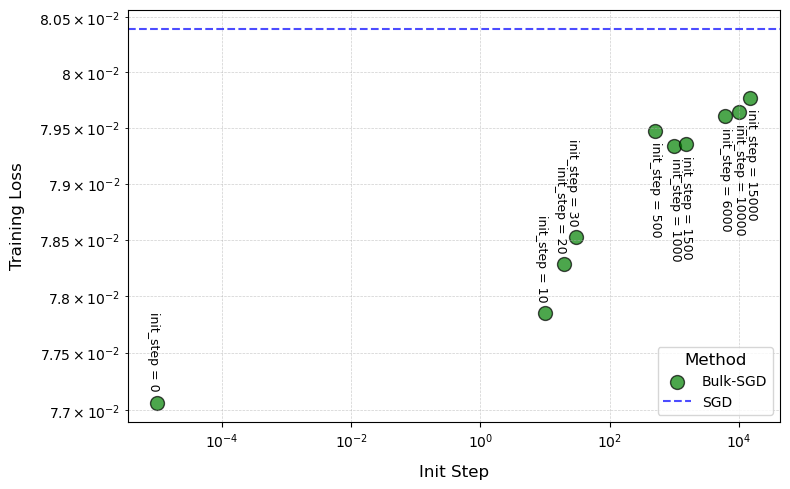}
        \caption{Training loss vs. switch point step}
        \label{fig:sub2}
    \end{subfigure}
    \hfill
    \begin{subfigure}[b]{0.32\linewidth}
        \centering
        \includegraphics[width=\linewidth]{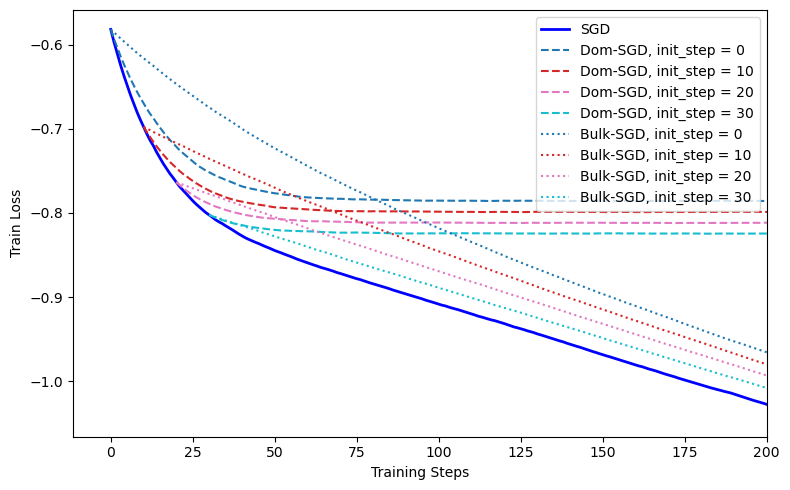}
        \caption{First 200 steps}
        \label{fig:sub3}
    \end{subfigure}
    \caption{Combined results from switch point experiments. Training was performed with an MLP on the MNIST-5k dataset.}
    \label{fig:combined}
\end{figure}

In the original study by \cite{song2025doessgdreallyhappen}, the \textit{Dom}-SGD and \textit{Bulk}-SGD algorithms are initialized at fixed points during training with SGD. We refer to these transition steps as \emph{switch points}. To evaluate the robustness of Phenomena~\ref{phenom:dom} and~\ref{phenom:bulk}, we investigate whether their conclusions hold across different switch points.

\textbf{Protocol.} We run baseline SGD training with checkpoints taken at various points along the optimization trajectory. At each switch point, we transition to either \textit{Dom}-SGD or \textit{Bulk}-SGD and continue training from that point onward. Aside from the earliest stages, our findings confirm that the observed behaviors remain consistent across different switch points.

\textbf{\textit{Dom}-SGD saturates early.} When initialized shortly after training begins (e.g., at step 30), \textit{Dom}-SGD continues to make progress only until around step 50. This suggests a rapid saturation of the information captured by the dominant subspace. In contrast, \textit{Bulk}-SGD trains more slowly during this early phase.

This observation aligns with ~\cite{gurari2018gradientdescenthappenstiny}, who show that gradient alignment with the dominant subspace emerges only after the initial steps of training.

\textbf{\textit{Bulk}-SGD achieves lower loss across switch points.} Plotting the final training loss of \textit{Bulk}-SGD for various switch points reveals a consistent pattern: \textit{Bulk}-SGD achieves lower training loss compared to SGD across most switch points. Notably, the lowest losses occur when switching early (at steps 0, 10, 20, or 30). 

Furthermore, we observe that \textit{Bulk}-SGD loss curves tend to be smoother than those of SGD. This suggests that removing the dominant component of the gradient may reduce stochastic fluctuations in the learning dynamics, possibly revealing a more stable optimization path.

\subsection{Instability of \textit{Bulk}-SGD at Larger Learning Rates}
\label{section:instability}

\begin{figure}[ht]
    \centering
    \begin{subfigure}[b]{0.45\textwidth}
        \centering
        \includegraphics[width=\textwidth]{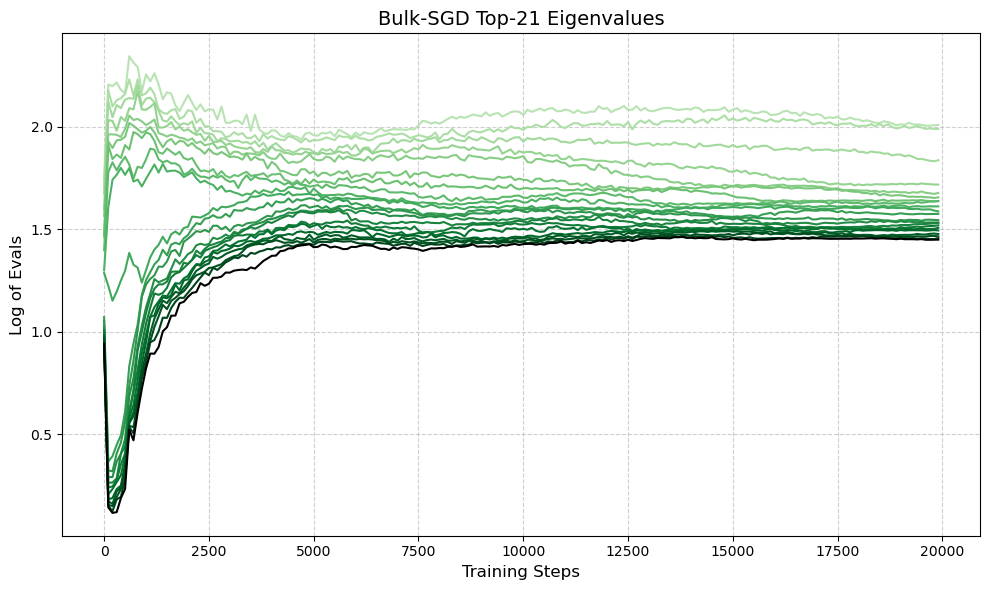}
        \caption{}
        \label{subfig:sgd_top_evals}
    \end{subfigure}
    \hfill
    \begin{subfigure}[b]{0.45\textwidth}
        \centering
        \includegraphics[width=\textwidth]{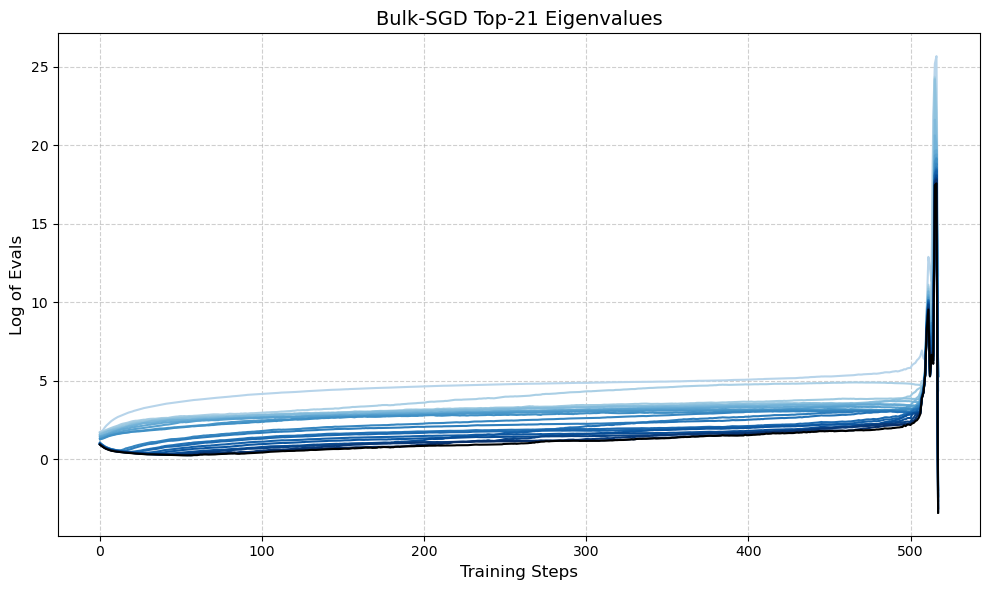}
        \caption{}
        \label{subfig:bulk_sgd_top_evals}
    \end{subfigure}
    \caption{Instability of \textit{Bulk}-SGD is caused by the sharpness explosion. Training was performed on MNIST-5k dataset. In \ref{subfig:sgd_top_evals} and \ref{subfig:bulk_sgd_top_evals}, we plot the 21 leading eigenvectors for SGD and Bulk-SGD respectively. For Bulk-SGD training Dominant subspace of dimension $k = 20$ was used. The 21 eigenvector is marked with black color.}
    \label{fig:instability_sharpness}
\end{figure}

While \textit{Bulk}-SGD demonstrates competitive performance in certain regimes, we observe that it becomes increasingly unstable when trained with larger learning rates. In particular, the training loss frequently diverges or spikes to large values. This instability is robust: it arises across different network architectures, activation functions, and both cross-entropy (CE) and mean squared error (MSE) loss functions. As such, it presents a key obstacle to using \textit{Bulk}-SGD for accelerated training.

A closer analysis of the optimization dynamics provides further insight. Under standard SGD, gradient norms typically decay over time as the parameters approach a local minimum. Surprisingly, in the case of \textit{Bulk}-SGD, we observe the opposite: gradient norms increase as training progresses.

\textbf{Gradient instability.} This anomalous behavior suggests that the exclusion of the dominant subspace alters the geometry of the optimization path in a way that destabilizes training.

\textbf{Sharpness and eigenvector dynamics.} We measure sharpness as the magnitude of the largest eigenvalues of the Hessian (See \ref{fig:instability_sharpness}). For SGD, eigenvalues initially rise before plateauing—a behavior that reflects stabilization of curvature in the loss landscape. In contrast, for \textit{Bulk}-SGD, sharpness steadily increases throughout training without saturation. Additionally, the leading Hessian eigenvectors grow progressively rather than stabilizing early, as is observed in SGD.

These trends suggest a link between increasing sharpness and gradient instability. Notably, the increase in sharpness is not limited to the dominant subspace—it is observed across the entire spectrum. This may explain the growing gradient norms and the ultimate failure of training under large learning rates.

\textbf{A stabilizing role of the dominant subspace.} These findings hint at an additional function of the dominant subspace: controlling sharpness. In agreement with ~\cite{song2025doessgdreallyhappen}, who observed that \textit{Dom}-SGD tends to reduce sharpness while \textit{Bulk}-SGD increases it, we posit that the dominant subspace plays an essential stabilizing role in the training dynamics of neural networks.
\begin{figure}[b]
    \centering
    \includegraphics[width=0.5\textwidth]{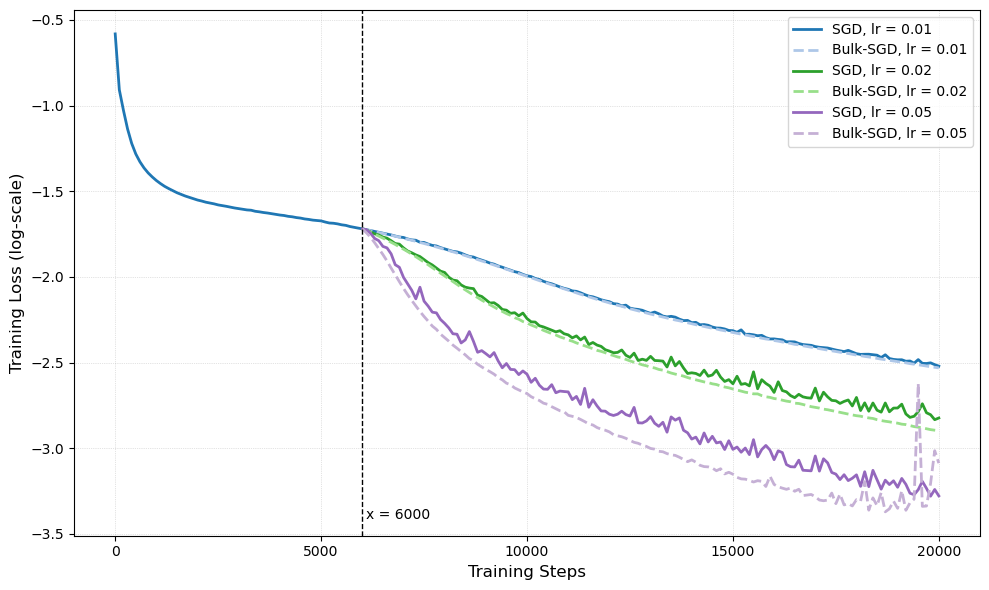}
    \caption{\textbf{Acceleration experiment}. SGD and \textit{Bulk}-SGD training loss on the MNIST-5k dataset. Each run is initialized after SGD 6{,}000 SGD steps.}
    \label{fig:acceleration_experiment}
\end{figure}

\section{Toward Accelerated Optimization via the Bulk Subspace}
\label{section:acceleration}

The observations from \cref{phenom:alignment,phenom:dom,phenom:bulk} point to a compelling interpretation of gradient dynamics in neural network training. Throughout most of the optimization trajectory, the parameters appear to \emph{follow the noise}—that is, gradients are predominantly aligned with the high-curvature \textit{Dominant subspace}. In contrast, the flatter \textit{Bulk} component, may carry the true learning signal.

The dominant subspace limits step sizes and introduces high variance. This leads us to hypothesize that restricting updates to the \textit{Bulk subspace} could reduce gradient noise and permit the use of larger learning rates. If successful, such a strategy could accelerate optimization and potentially reach lower minima.

\textbf{Experimental setup.} We test this hypothesis by training a neural network for 6{,}000 steps, switching to \textit{Bulk}-SGD partway through training. After the switch, we optionally increase the learning rate to exploit the flatter curvature directions in the \textit{Bulk subspace}. The goal is to isolate the learning signal from stochastic noise and assess whether optimization can proceed more effectively.

\textbf{Results.} As shown in \cref{fig:acceleration_experiment}, switching to \textit{Bulk}-SGD with a larger learning rate leads to an immediate acceleration in training: the loss decreases faster compared to standard SGD under the same rate. However, in the later stages of training, instability in the training loss emerges, mirroring earlier findings on \textit{Bulk}-SGD dynamics.

\begin{phenomenon}
\label{phenom:acceleration_vs_stability}
\textbf{Faster descent, fragile stability.} \textit{Bulk}-SGD supports faster initial optimization at higher learning rates, but eventually suffers from instability—highlighting the stabilizing role of the \textit{Dominant subspace}.
\end{phenomenon}

These findings lend partial support to our hypothesis: restricting updates to the \textit{Bulk subspace} can reduce noise and allow for more aggressive learning rates—but only temporarily. The eventual breakdown underscores the dual role of the \textit{Dominant} component in stabilizing curvature.

In the next section, we explore a hybrid approach that interpolates between \textit{Bulk}-SGD and \textit{Dom}-SGD, aiming to combine their complementary advantages.

\subsection{Interpolated Steps}

\begin{figure}[htbp]
    \centering
    \begin{subfigure}[b]{0.45\textwidth}
        \centering
        \includegraphics[width=\textwidth]{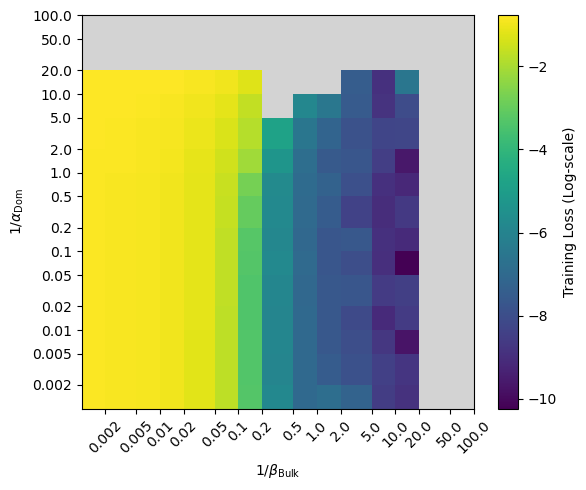}
        \caption{}
        \label{subfig:interpolated_expoeriments_train_loss_log}
    \end{subfigure}
    \hfill
    \begin{subfigure}[b]{0.45\textwidth}
        \centering
        \includegraphics[width=\textwidth]{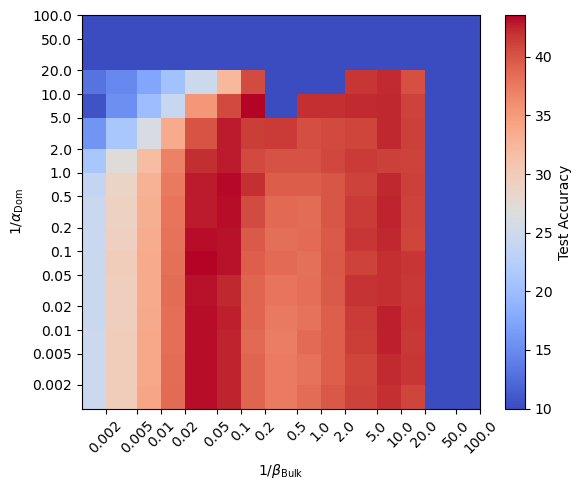}
        \caption{}
        \label{subfig:interpolated_expoeriments_test_acc}
    \end{subfigure}
    \caption{Training neural network using interpolated gradient steps. Heatmaps are formed by training with different parameters $\alpha_{\mathrm{Dom}}$ and $\beta_{\mathrm{Bulk}}$ on CIFAR10-5k dataset. In \ref{subfig:interpolated_expoeriments_train_loss_log} we plot the training loss and in \ref{subfig:interpolated_expoeriments_test_acc} we plot the test accuracy. Grey cells indicate training setups that resulted in the diverging loss. For more detailed version see \cref{fig:interpolation_experiment_test_acc_full,fig:interpolation_experiment_training_loss_full}.} 
    \label{fig:interpolated_experiments}
\end{figure}

To harness the complementary benefits of the \textit{Bulk} and \textit{Dominant subspaces}—namely, the accelerated descent along flat directions and the stabilizing effect on sharpness—we propose an interpolated update rule that combines their respective contributions. Specifically, we compute the gradient $g_t = \nabla\mathcal{L}(\theta_t)$ and update the parameters as follows:

\[
\theta_{t+1} \leftarrow \theta_t 
- \frac{\eta}{\alpha_{\mathrm{Dom}}} P_k(\theta_t) g_t
- \frac{\eta}{\beta_{\mathrm{Bulk}}} P_k^{\perp}(\theta_t) g_t,
\]

where $P_k$ projects onto the top-$k$ eigenvectors of the Hessian (the \textit{Dominant subspace}), and $P_k^\perp$ denotes the projection onto its orthogonal complement (the \textit{Bulk subspace}).

This formulation smoothly interpolates between different gradient descent regimes:
\begin{itemize}
    \item When $\alpha_{\mathrm{Dom}} = \beta_{\mathrm{Bulk}}$, the update reduces to standard gradient descent.
    \item In the limit $\alpha_{\mathrm{Dom}} \rightarrow \infty$, the dominant component is suppressed, recovering \textit{Bulk}-SGD.
    \item Conversely, $\beta_{\mathrm{Bulk}} \rightarrow \infty$ recovers \textit{Dom}-SGD.
\end{itemize}

\textbf{Results.} In \Cref{fig:interpolated_experiments}, we visualize the training loss and test accuracy across a range of $\alpha_{\mathrm{Dom}}$ and $\beta_{\mathrm{Bulk}}$ values. We observe that lower training losses are consistently associated with higher contributions from the \textit{Bulk subspace} (i.e., larger $\frac{1}{\beta_{\mathrm{Bulk}}}$). This reinforces our earlier finding that promoting the \textit{Bulk} component enables faster descent and helps locate lower minima.

However, this advantage does not straightforwardly translate into improved test accuracy. The highest test accuracies are achieved when the \textit{Dominant subspace} is relatively more emphasized—i.e., when $\frac{1}{\alpha_{\mathrm{Dom}}}$ is large compared to $\frac{1}{\beta_{\mathrm{Bulk}}}$. These results suggest a nuanced trade-off:

\begin{phenomenon}
\label{phenom:acceleration_vs_generalization}
\textbf{Bulk accelerates, Dom generalizes.} Interpolated updates reveal that \textit{Bulk}-dominated steps improve training loss, while \textit{Dom}-oriented steps support generalization—highlighting a tension between optimization speed and test-time robustness.
\end{phenomenon}

Interestingly, we also find that competitive test performance can emerge from both \textit{Bulk}- and \textit{Dom}-promoting regimes, depending on the interpolation balance. This points to the possibility of a principled hybrid approach that adapts the interpolation weights throughout training to exploit early acceleration while preserving generalization.

\section{Bulk Subspaces and Hessian Decompositions}
\label{section:hessian_decomp}

\begin{figure}[ht]
    \centering
    \begin{subfigure}[b]{0.3\textwidth}
        \centering
        \includegraphics[width=\textwidth]{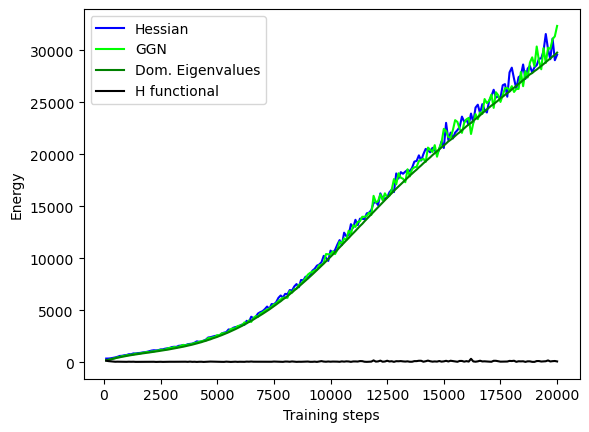}
        \caption{Total matrix energy}
    \end{subfigure}
    \hfill
    \begin{subfigure}[b]{0.3\textwidth}
        \centering
        \includegraphics[width=\textwidth]{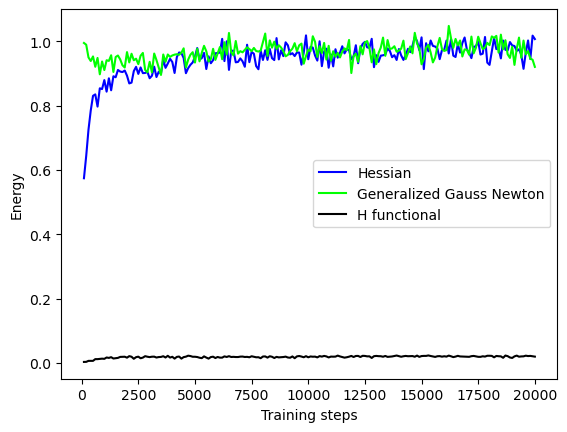}
        \caption{$\norm{M}_{S_\text{Dom}}^2 / \norm{M}_{F}^2$}
        \label{subfig:mse_tanh_m_energy_ratio}
    \end{subfigure}
    \hfill
    \begin{subfigure}[b]{0.3\textwidth}
        \centering
        \includegraphics[width=\textwidth]{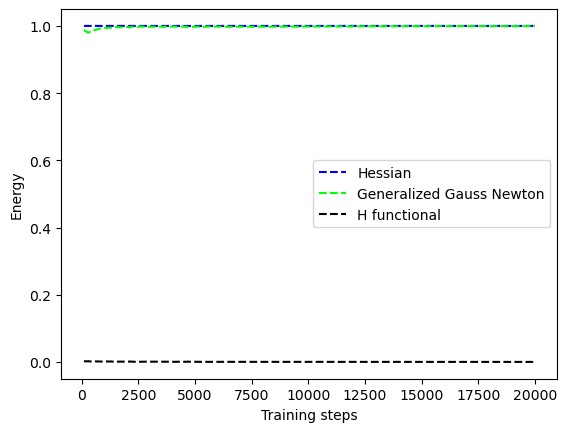}
        \caption{$\norm{M}_{S_\text{Dom}}^2 / \norm{S_{\text{Dom}}}^2$}
        \label{subfig:mse_tanh_subspace_energy_ratio}
    \end{subfigure}
    \caption{Hessian energy analysis. MSE Loss and Tanh activation. The rest of the parameters are the same as in Fig. }
    \label{fig:hessian_energy_mse_tanh}
\end{figure}

\begin{figure}[ht]
    \centering
    \begin{subfigure}[b]{0.3\textwidth}
        \centering
        \includegraphics[width=\textwidth]{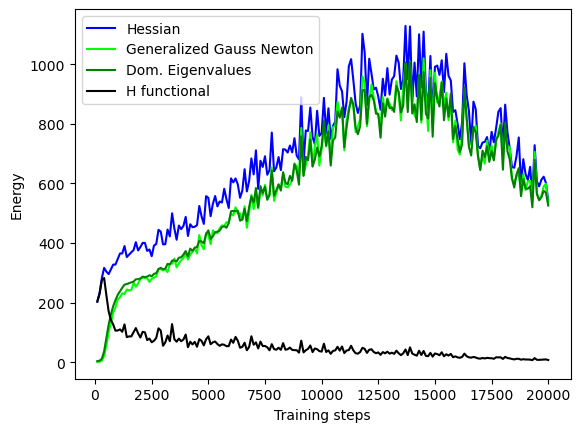}
        \caption{Total matrix energy}
    \end{subfigure}
    \hfill
    \begin{subfigure}[b]{0.3\textwidth}
        \centering
        \includegraphics[width=\textwidth]{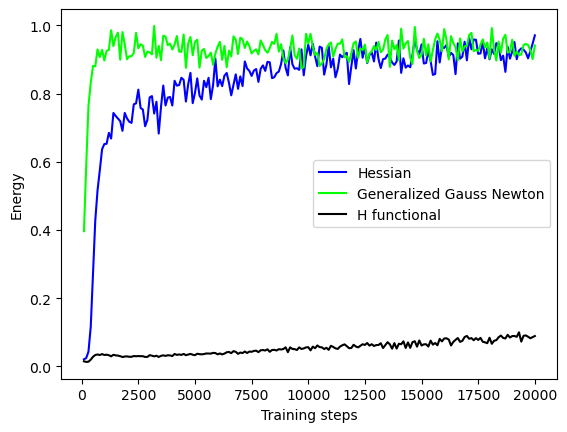}
        \caption{$\norm{M}_{S_\text{Dom}}^2 / \norm{M}_{F}^2$}
        \label{subfig:ce_tanh_m_energy_ratio}
    \end{subfigure}
    \hfill
    \begin{subfigure}[b]{0.3\textwidth}
        \centering
        \includegraphics[width=\textwidth]{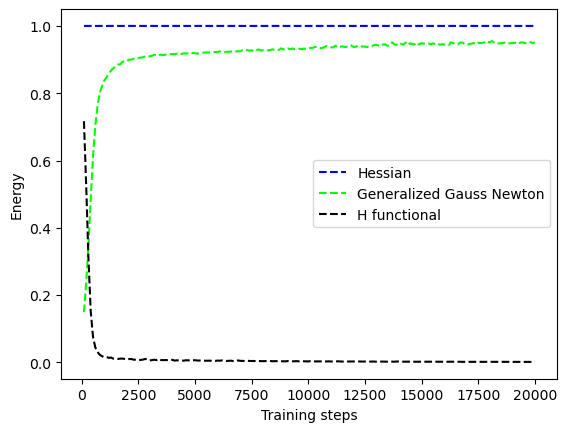}
        \caption{$\norm{M}_{S_\text{Dom}}^2 / \norm{S_{\text{Dom}}}^2$}
        \label{subfig:ce_tanh_subspace_energy_ratio}
    \end{subfigure}
    \caption{Hessian energy analysis. Cross-Entropy Loss and Tanh activation. The rest of the parameters are the same as in Fig.}
    \label{fig:hessian_energy_ce_tanh}
\end{figure}

\begin{figure}[ht]
    \centering
    \begin{subfigure}[b]{0.3\textwidth}
        \centering
        \includegraphics[width=\textwidth]{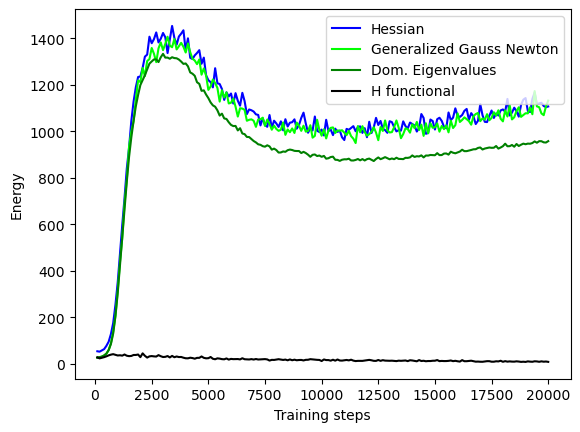}
        \caption{Total matrix energy}
    \end{subfigure}
    \hfill
    \begin{subfigure}[b]{0.3\textwidth}
        \centering
        \includegraphics[width=\textwidth]{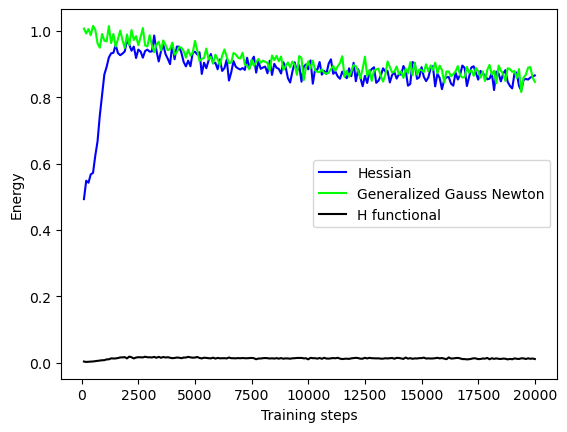}
        \caption{$\norm{M}_{S_\text{Dom}}^2 / \norm{M}_{F}^2$}
        \label{subfig:mse_relu_m_energy_ratio}
    \end{subfigure}
    \hfill
    \begin{subfigure}[b]{0.3\textwidth}
        \centering
        \includegraphics[width=\textwidth]{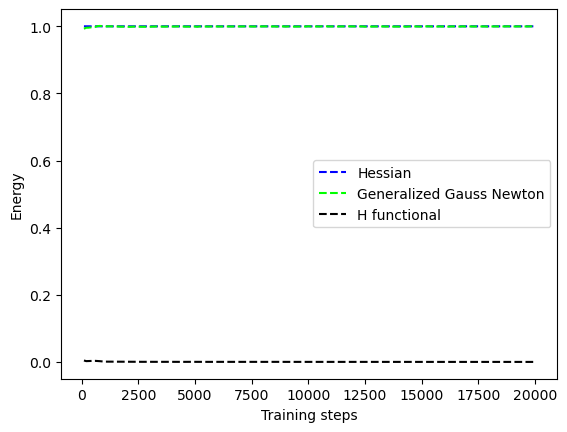}
        \caption{$\norm{M}_{S_\text{Dom}}^2 / \norm{S_{\text{Dom}}}^2$}
        \label{subfig:mse_relu_subspace_energy_ratio}
    \end{subfigure}
    \caption{Hessian energy analysis. MSE Loss and ReLU activation. The rest of the parameters are the same as in Fig. }
    \label{fig:hessian_energy_mse_relu}
\end{figure}

The works \cite{song2025doessgdreallyhappen} and \cite{gurari2018gradientdescenthappenstiny} provide primarily empirical insights into the structure and behavior of the \textit{Bulk subspace}. However, a complete theoretical framework capable of capturing these phenomena remains an open area of investigation. An improved understanding of the \textit{Bulk} could potentially inspire new optimization algorithms that exploit its flatness to accelerate training.

One promising direction is to study the structure of the Hessian via its decomposition into two components:
\begin{equation}
    H_{\mathcal{L}} = H_o + H_f,
\end{equation}
where
\begin{align}
    H_o &= \mathbb{E}_{p(x, y)}\left[ \nabla_\theta F(x)^\top \left( \nabla^2 l_{x, y} \right) \nabla_\theta F(x) \right], \\
    H_f &= \mathbb{E}_{p(x, y)}\left[ \sum_{c=1}^{K} \left( \partial l_{x, y} \right)_c \nabla^2_\theta F_c(x) \right].
\end{align}

The term $H_o$ is known as the \textit{outer-product Hessian}, or the \textit{Generalized Gauss–Newton} matrix. This decomposition follows naturally from the application of the chain rule. Intuitively, $H_o$ captures the curvature induced by the linearized model, whereas $H_f$, also referred to as the \textit{functional Hessian}  (\cite{singh2021analyticinsightsstructurerank,singh2023hessian}), encodes curvature stemming from second-order derivatives of the model outputs with respect to parameters.

To analyze the contribution of each component to training dynamics, we define the \textbf{Hessian energy} as the squared Frobenius norm:
\[
\norm{H}_F^2 = \sum_{i,j} H_{ij}^2.
\]
This quantity reflects the total curvature magnitude in parameter space. Similarly, we compute $\norm{H_o}_F^2$ and $\norm{H_f}_F^2$ to measure the respective energies of the outer-product and functional Hessians.

We are particularly interested in how curvature is distributed along the \textit{Dominant subspace}. Let $S_k = \Span{v_1, v_2, \dots, v_k}$ denote the top-$k$ eigenvectors of the Hessian. We define the energy of $H$ restricted to this subspace as:
\[
\norm{H}_{S_k}^2 = \sum_{i=1}^k v_i^\top H^2 v_i.
\]
As shown in \Cref{fig:hessian_energy_mse_tanh,fig:hessian_energy_ce_tanh,fig:hessian_energy_mse_relu}, the energies of $H$ and $H_o$ are significantly larger than that of $H_f$, consistent with the common practice of approximating $H$ by $H_o$.

An alternative view expresses the Dominant subspace energy via the top eigenvalues of $H$:
\[
\norm{S_k}^2 = \sum_{i=1}^k \lambda_i^2,
\]
where $\lambda_1, \dots, \lambda_k$ are the leading eigenvalues. This formulation highlights that the Dominant subspace captures a significant portion of total Hessian energy and that this energy is well-aligned with that of the outer-product Hessian.

To compare the alignment of $H_o$ and $H_f$ with the Dominant subspace, we define:
\[
\norm{H_o}_{S_k}^2 = \sum_{i=1}^k v_i^\top H_o^2 v_i, \quad 
\norm{H_f}_{S_k}^2 = \sum_{i=1}^k v_i^\top H_f^2 v_i.
\]

In \cref{subfig:mse_tanh_m_energy_ratio,subfig:ce_tanh_m_energy_ratio,subfig:mse_relu_m_energy_ratio}, we observe that, after early training, the energies of $H$ and $H_o$ concentrate within the Dominant subspace. Likewise, in \cref{subfig:mse_tanh_subspace_energy_ratio,subfig:ce_tanh_subspace_energy_ratio,subfig:mse_relu_subspace_energy_ratio}, we find that the Dominant subspace energy is dominated by $H_o$, with negligible contribution from $H_f$.

These observations suggest that while $H_o$ governs the principal directions of sharp curvature, $H_f$ is largely unaligned with these directions. This supports the hypothesis that the \textit{Bulk subspace} may be shaped more by the residual structure of $H_f$, and motivates further study of the interactions between curvature, sharpness, and optimization behavior in the \textit{Bulk} and \textit{Dominant subspaces}.

\section{Conclusion}

In this work, we build upon the studies of \cite{gurari2018gradientdescenthappenstiny} and \cite{song2025doessgdreallyhappen}, and through a series of ablation experiments, we draw the following conclusions: 

\begin{itemize}
    \item \textit{Bulk}-SGD can accelerate neural network training, leading to lower training loss. However, its performance is sensitive to key hyperparameters and suffers from instability.
    \item The \textit{Dominant} component of the gradient plays a critical role in ensuring stability and promoting generalization during training.
\end{itemize} 

\textbf{Next steps.} A primary challenge for practical adoption is the instability observed in \textit{Bulk}-SGD. In future work, we plan to conduct further experiments aimed at addressing this issue, exploring techniques such as weight decay, normalization, and adaptive scheduling.

Additionally, many theoretical questions remain open regarding the precise roles and interactions of the phenomena described in this paper. A deeper understanding of these dynamics will be crucial for refining and extending curvature-aware optimization methods.

\bibliographystyle{apalike}
\bibliography{references} 

\appendix
\section{Experimental Setup}

The architecture and training setup were chosen following \cite{song2025doessgdreallyhappen}.  
Two architectures were used in the experiments:  
\begin{itemize}
    \item \textbf{MLP}: A 3-layer fully connected neural network with 100 hidden units per layer.
    \item \textbf{CNN}: A 3-layer convolutional neural network with 32 channels per convolutional layer.
\end{itemize}

In all experiments, a batch size of 50 was used, and training was performed for 200 epochs using the mean squared error (MSE) loss. We use the MNIST-5k and CIFAR10-5k datasets—5{,}000-sample subsets of the MNIST \cite{mnist} and CIFAR-10 \cite{cifar10} datasets, respectively. Unless otherwise specified, the number of classes $k=10$ was chosen as the dimension of the \textit{Dominant subspace}.

To reduce training time, the update frequency for the Dominant subspace was reduced as follows:
\begin{itemize}
\item In \cref{table:ablation_mlp,table:ablation_cnn}, updated every 10 steps.
\item In \cref{fig:interpolated_experiments}, updated every 100 steps.
\end{itemize}

\section{Full Interpolation Experiment Results}

\begin{figure}[ht]
    \centering
    \includegraphics[width=0.5\linewidth]{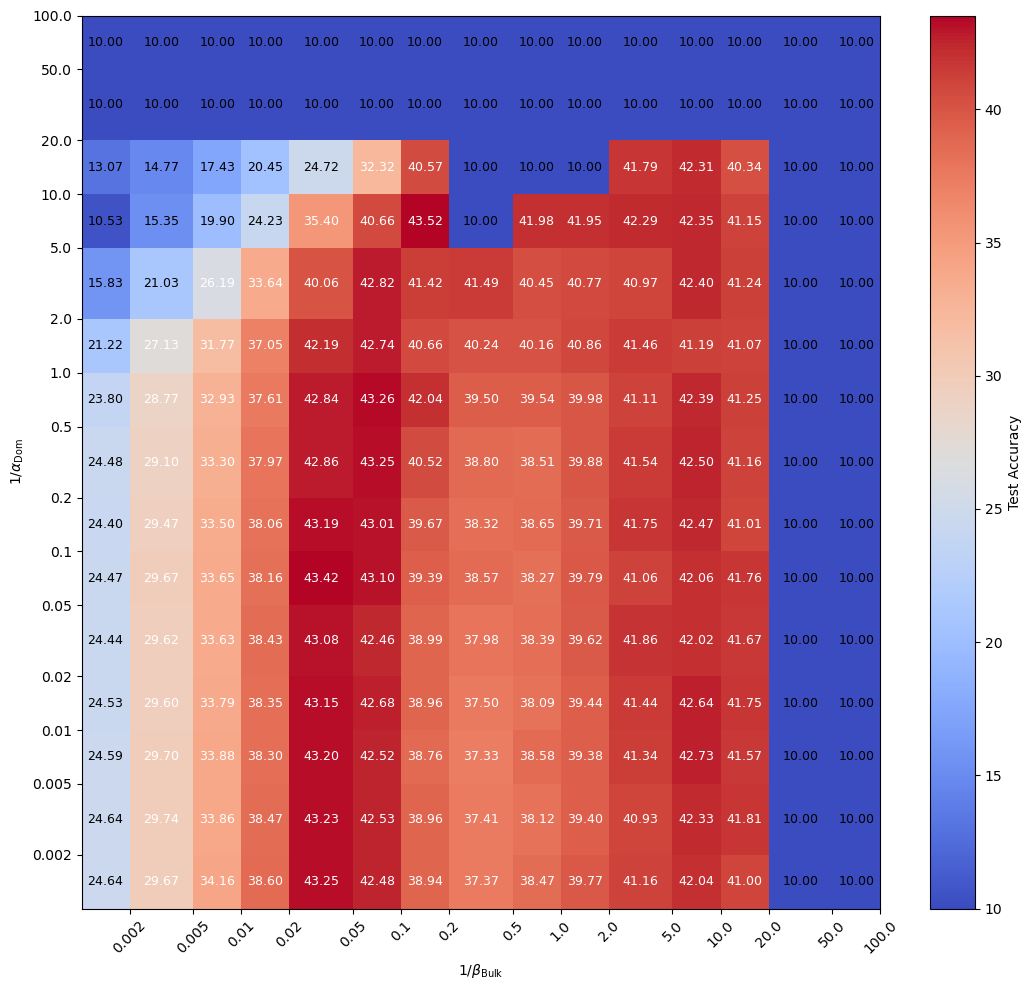}
    \caption{Training loss for the neural network trained with the interpolated gradient steps. }
    \label{fig:interpolation_experiment_test_acc_full}
\end{figure}

\begin{figure}[ht]
    \centering
    \includegraphics[width=0.5\linewidth]{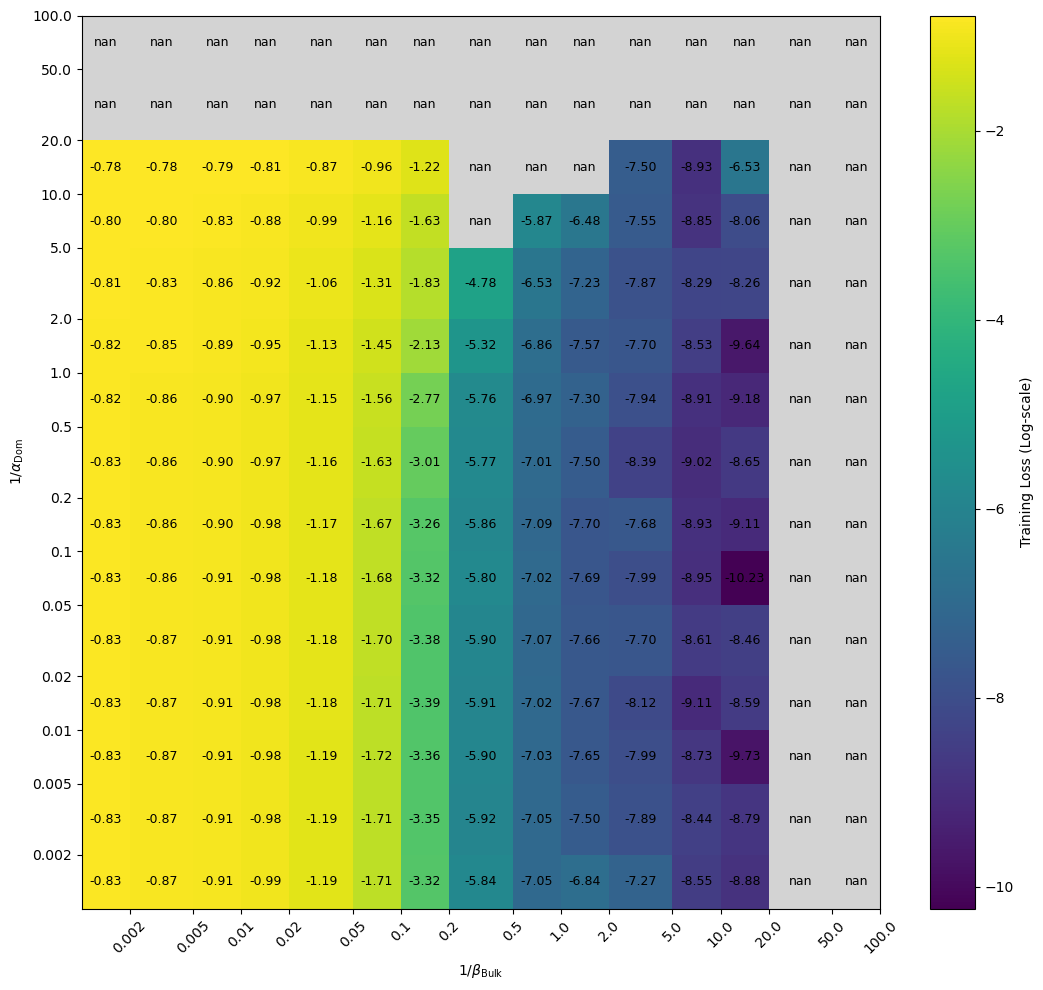}
    \caption{Test accuracy for the neural network trained with the interpolated gradient steps.}
    \label{fig:interpolation_experiment_training_loss_full}
\end{figure}

You can find the full results with the numerical values for training loss and test accuracy in \cref{fig:interpolation_experiment_test_acc_full, fig:interpolation_experiment_training_loss_full}.

\end{document}